\def\year{2022}\relax
\documentclass[letterpaper]{article} 
\usepackage{aaai22}  
\usepackage{times}  
\usepackage{helvet}  
\usepackage{courier}  
\usepackage[hyphens]{url}  
\usepackage{graphicx} 
\usepackage{natbib}  
\usepackage{caption} 
\DeclareCaptionStyle{ruled}{labelfont=normalfont,labelsep=colon,strut=off} 
\usepackage{algorithm}
\usepackage{algorithmic}
\usepackage{amsfonts}

\usepackage{todonotes}

\usepackage{booktabs}

\usepackage[T1]{fontenc}    

\usepackage{newfloat}
\usepackage{listings}
\floatstyle{ruled}
\newfloat{listing}{tb}{lst}{}
\floatname{listing}{Listing}
\title{Improving and Diagnosing Knowledge-Based Visual Question Answering via Entity Enhanced Knowledge Injection}
\author{
    Diego Garcia-Olano, Yasumasa Onoe, Joydeep Ghosh
}
\affiliations{
    University of Texas At Austin\\

    diegoolano@utexas.edu, yasumasa@utexas.edu, jghosh@utexas.edu
}

\begin{document}

\maketitle

\begin{abstract}
Knowledge-Based Visual Question Answering (KBVQA) is a bi-modal task requiring external world knowledge in order to correctly answer a text question and associated image. Recent single modality text work has shown knowledge injection into pre-trained language models, specifically entity enhanced knowledge graph embeddings, can improve performance on downstream entity-centric tasks.  In this work, we empirically study how and whether such methods, applied in a bi-modal setting, can improve an existing VQA system's performance on the KBVQA task.  We experiment with two large publicly available VQA datasets, (1) KVQA which contains mostly rare Wikipedia entities and (2) OKVQA which is less entity-centric and more aligned with common sense reasoning.  Both lack explicit entity spans and we study the effect of different weakly supervised and manual methods for obtaining them. Additionally we analyze how recently proposed bi-modal and single modal attention explanations are affected by the incorporation of such entity enhanced representations.  Our results show substantial improved performance on the KBVQA task without the need for additional costly pre-training and we provide insights for when entity knowledge injection helps improve a model's understanding.   
We provide code and enhanced datasets for reproducibility.
\end{abstract}

\section{Introduction}

Visual Question Answering (VQA) is a multi-modal task that involves correctly answering a text question pertaining to an associated image without the explicit need for external world knowledge (facts about history, geography, etc). In addition to datasets involving the need for commonsense reasoning \citep{okvqa, fvqa}, recent work on \textit{knowledge-based VQA} \citep{shahMYP19} involve questions whose answers explicitly require external knowledge about named entities within an image. 
Common amongst state of the art VQA solutions \citep{singh2020mmf} is the need for large amounts of computational resources and supervised question-image pairs in order to pretrain models that generalize well.  Recent work E-BERT \citep{poerner-etal-2020-e} gives improved performance on single modality, entity-centric text tasks by using efficient external knowledge injection into pre-trained Transformer language models (LMs).  Although there has been quite a bit of work studying whether LMs can be used as knowledge bases\cite{Fabio_Petroni_19, Adam_Roberts_20, Nina_Poerner_19} there has been less attention on how this affects vision-language models.  Additionally, while research on interpretability methods for single modalities is abundant \citep{integratedgradients, shapvalues, tracin}, saliency maps for images or feature attribution methods for text for instance, only recently are there methods explicitly targeted for bi-modal tasks like VQA, namely the bi-modal generic attention explainability method BM-GAE \citep{BMGAE} which provides a promising method by which to understand image and text explanations jointly. \\
\indent In this work we analyze how knowledge injection via E-BERT affects the performance on an existing visual-linguistic model LXMERT \citep{tan-bansal-2019-lxmert} on the relatively unexplored task of knowledge-based VQA (KBVQA) both in terms of accuracy and explainability via BM-GAE.  
\indent We experiment using two large publicly available VQA datasets: (i)KVQA \citep{shahMYP19} that is explicitly tied to Wikipedia and rich in rare entities and (ii) OKVQA \citep{okvqa} that is less entity-centric and more aligned with common sense reasoning. Both datasets lack explicit entity spans and we show how using different entity sets resulting from either weakly supervised methods or manual human annotation affects knowledge injection on task performance. \\
\indent Our analysis shows improved performance on the entity rich KVQA dataset, 2.5\% top 1 accuracy, and a smaller improvement on the OKVQA dataset.  Both come without the need for any additional costly pre-training; for a given dataset, simply fine tune LXMERT using knowledge injection via E-BERT and do inference on its test set.  In addition to error analysis, we assess the effect of E-BERT on the explanations generated by BM-GAE and provide insights for when entity injection helps improve a model's understanding for knowledge aware VQA. Our weakly and manually enhanced datasets and code are made available at  \texttt{anonymous.4open.science/r/kbvqa\_xai/}.


\section{Background and Related Work}
In this section we provide some background on E-BERT, bi-modal generic attention explainability for VQA and LXMERT along with related works.

\paragraph{E-BERT} Wikipedia2Vec \cite{yamada-etal-2016-joint} embeds
words and entities (Wikipedia URLs) into a common space.  Given a vocabulary of words $\mathbb{L}_{\mathrm{Word}}$
and a vocabulary of entities $\mathbb{L}_{\mathrm{Ent}}$, it learns a lookup
embedding function $\mathcal{E}_{\mathrm{Wikipedia}} : \mathbb{L}_{\mathrm{Word}} \cup \mathbb{L}_{\mathrm{Ent}} \rightarrow R^{d_{\mathrm{Wikipedia}}}$ .The E-BERT authors \cite{poerner-etal-2020-e} align Wikipedia2Vec entity vectors $\mathcal{E}_{\mathrm{Wikipedia}}[\mathbb{L}_{\mathrm{Ent}}]$ with BERT’s wordpiece vector space $\mathcal{E}_{\mathrm{BERT}}[\mathbb{L}_{\mathrm{WP}}]$ and then feed these aligned vectors into BERT as if they came from BERT's native wordpiece space. This procedure allows E-BERT to inject knowledge into BERT without making any changes to the BERT encoder itself or doing any additional pretraining.    Specifically given an $x \in \mathbb{L}_{\mathrm{WP}} \cap \mathbb{L}_{\mathrm{Word}}$ they learn an unconstrained linear mapping $\mathbf{W} \in  \mathbb{R}^{d_{\mathrm{BERT}} \times d_{\mathrm{Wikipedia}}}$  that seeks to minimize \begin{equation}
 \label{loss1}
 \sum_{x \in \mathbb{L}_{\mathrm{WP}} \cap \mathbb{L}_{\mathrm{Word}}} || \mathbf{W}{\mathcal{E}}_{\mathrm{Wikipedia}}(x) - {\mathcal{E}}_{\mathrm{BERT}}(x) {||}_{2}^{2}
 \end{equation}
Since Wikipedia2Vec embeds $\mathbb{L}_{\mathrm{Word}}$ and $\mathbb{L}_{\mathrm{Ent}}$ into
the same space, $W$ can be applied
to $L_{\mathrm{Ent}}$ as well.  Thus at inference time they simply use $\mathbf{W}$ to construct an ${\mathcal{E}}_{\mathrm{E-BERT}}$ entity embeddings via ${\mathcal{E}}_{\mathrm{E-BERT}}(a) = {\mathbf{W}{E}_{\mathrm{Wikipedia}}(a)}$ for an entity $a$ where $\mathcal{E}_{\mathrm{E-BERT}} : \mathbb{L}_{\mathrm{Ent}}  \rightarrow R^{d_{\mathrm{BERT}}}$ . They then prepend ${\mathcal{E}}_{\mathrm{E-BERT}}(a)$ to the BERT embedding of $a$ ( with a slash ``/" between them ).  They finally feed this updated input directly into their task classifier, and show improved accuracy and robustness measures for QA and other tasks.

\paragraph{Learning Cross-Modality Encoder Representations from Transformers} LXMERT \cite{tan-bansal-2019-lxmert} is a large-scale Transformer vision-language model consisting of three encoders: a visual object relationship encoder that leverages Fast R-CNN features, a language encoder (BERT base), and a cross-modality encoder that incorporates the prior visual and language encoder features.  The vision and language encoders leverage many self attention layers while the final cross modality uses co-attention over both modalities. It is pre-trained with large amounts of image-and-sentence pairs from 5 vision-language datasets via five diverse representative pre-training tasks: masked language modeling, masked object prediction (feature regression and label classification), cross-modality matching, and image question answering.

\paragraph{Bi-modal Generic Attention and VQA Explainations} In a recent work\citep{BMGAE} the authors propose BM-GAE, the first method to explain predictions by any Transformer-based architecture, particularly bi-modal Transformers that utilize co-attention like LXMERT, with word token feature importance and visual region saliency maps.  They show BM-GAE to be superior to all existing methods which are adapted from single modality methods via perturbation tests; ie, they identify important text tokens/regions via BM-GAE on the inference set and show removing the most important of them and re-doing inference has the most negative impact on accuracy compared with other methods.  The method uses the model’s attention layers to produce relevancy maps for each of the interactions between the input modalities in the network and is a generalization of TRF \citep{Chefer_2021_CVPR} without Layer-wise Relevance Propagation \citep{LRP}  which itself was shown to be effective on single-modality Transformers that utilize self-attention such as VilBERT\citep{lu2019vilbert}.  

Additional VQA explainability methods which focus on models that generate rationales for predicted answers \cite{ li-etal-2018-tell, jialinaaai} or as part of the inner process of a model \cite{nagaraj-rao-etal-2021-first} are of interest and can provide faithful explanations at the expense of additional input data supervision.  For this work we focus on methods which can be used with LXMERT as is without needing to re-train LXMERT with additional loss objectives or train auxiliary models to generate rationales that could be used in conjunction with LXMERT.

\paragraph{Knowledge-Based VQA} While there are quite a bit of methods for VQA and VQA requiring commonsense reasoning \cite{yu2020ernievil, Gan0LZ0020, lu2019vilbert, Su2020VL-BERT:, messina2020finegrained, shi2020contrastive, gao2019graphs, sampat2020visuo-linguistic}, there are fewer knowledge-based VQA ones \cite{BoostingVQA, Singh_2019_ICCV, garderes-etal-2020-conceptbert, ziaeefard-lecue-2020-towards, song2020kvlbert, shevchenko2021reasoning} and they follow a pattern of using visual, text and knowledge base embeddings separately or jointly for learning to produce answers. The two works most similar to ours either inject knowledge as a separate input into the cross attention module of LXMERT ( changing it to allow interaction with the text and a knowledge graph embedding input) \cite{garderes-etal-2020-conceptbert} or as an additional pre-training step that both identifies entity spans in a question and adds an additional objective pushing those knowledge graph representations towards their BERT entity representations \cite{shevchenko2021reasoning}.  Our work differs from these in that both of their solutions require pretraining of LXMERT whereas our proposal can be simply plugged in and used with LXMERT's existing pretrained weights and fine tuning for downstream tasks.

\begin{figure*}[t]
    \centering
    \includegraphics[width=.96\linewidth]{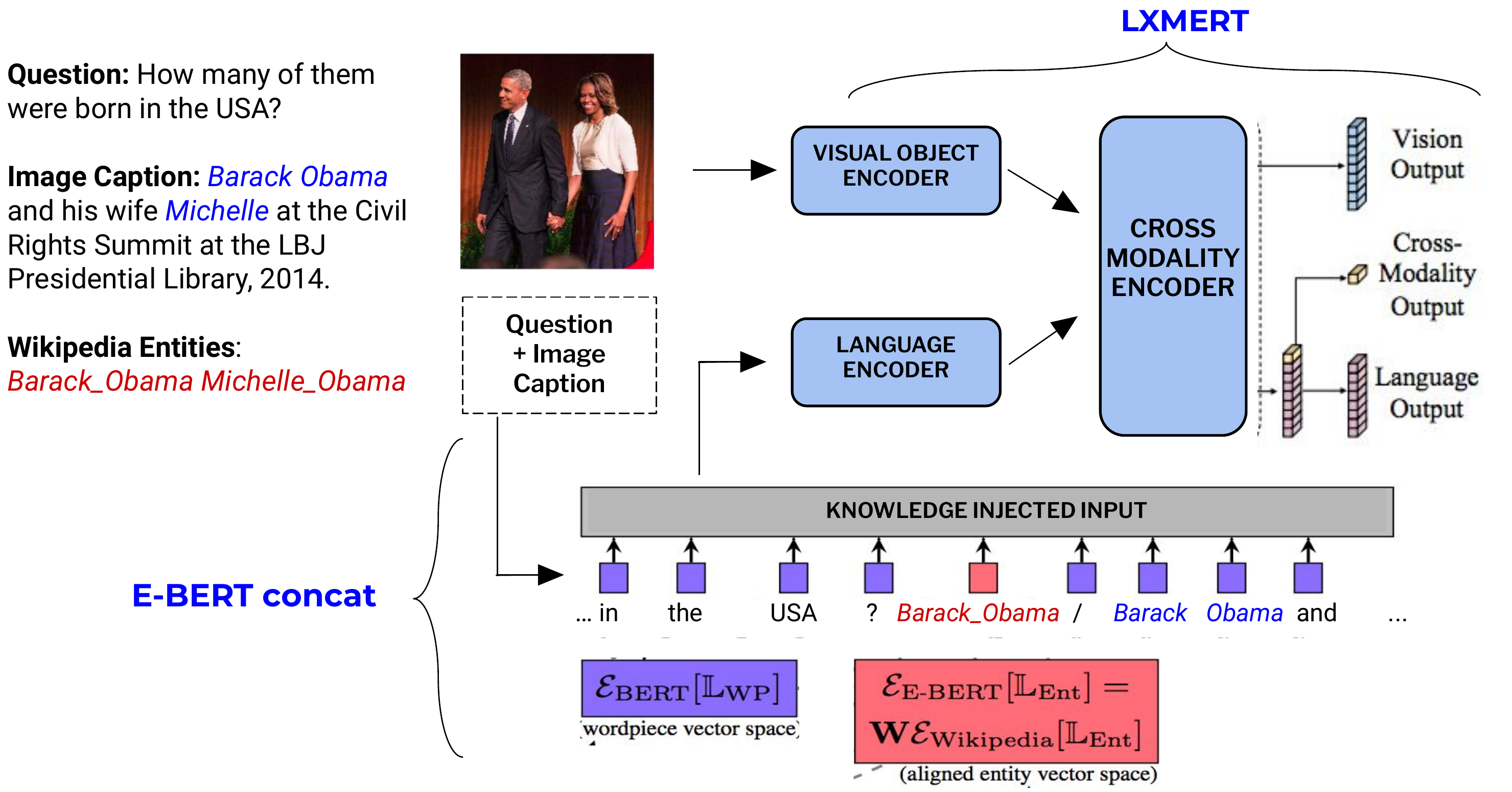}
    \caption{E-BERT knowledge injection into pre-trained LXMERT for knowledge aware VQA during fine-tuning. The example above is from the KVQA dataset where image captions are provided.  The use of such optional captions is studied here.}
    \label{fig:sys_arch_v2}
\end{figure*}

\section{Knowledge Injection via E-BERT for KBVQA}
Figure \ref{fig:sys_arch_v2} shows the manner in which we utilize E-BERT for knowledge injection into LXMERT's BERT model for knowledge aware VQA.  We first need to learn to map the 5.8 million Wikipedia entity embedding matrix the authors provide\footnote{https://wikipedia2vec.github.io/wikipedia2vec/} to that of the pre-trained LXMERT BERT space.  We utilize the code available in \cite{poerner-etal-2020-e} to learn that linear mapping ${\mathcal{E}}_{\mathrm{E-BERT}}$.
We then adapt LXMERT \footnote{https://github.com/airsplay/lxmert} to add E-BERT representations for each entity span in a given input question sequence during the tokenization phase as follows: We first check if an entity span $a$ exists in the WikipediaVec matrix via a direct map lookup and if so, generate the entity wikipedia2vec embedding using ${\mathcal{E}}_{\mathrm{E-BERT}}(a)$ to map that to the LXMERT BERT space.  We  finally append a BERT slash embedding followed by the BERT embedding of the entity $a$ to ${\mathcal{E}}_{\mathrm{E-BERT}}(a)$.  As an example if ``Barack Obama" is an entity that is found in the WikipediaVec matrix, the output for it while creating embeddings becomes ${\mathcal{E}}_{\mathrm{E-BERT}}$(``Barack Obama")) + BERT(``/") + BERT(``Barack Obama").  As WikipediaVec was trained using a cased tokenizer and LXMERT BERT is uncased by default, we need to titlecase the entity input string to WikipediaVec before hand as well. 

In the case that entity sets are not explicitly given with a downstream knowledge aware VQA task, we propose and study the important effect different methods for obtaining them affect downstream task performance. In the following section we discuss the datasets we use for our experiments and the methods we utilize to extract entity spans.

\section{Experiments}
\paragraph{KVQA dataset} The KVQA dataset \cite{shahMYP19} contains 24K images with text captions, 183K image/question QA pairs over 5 splits of the data ( median 7 questions per image), and associated metadata for the 18.8K unique Wikipedia entities displayed in those images (QID and Wikipage title).  The dataset has a large amount of QA pairs compared with other VQA datasets and its explicit Wiki supervision is unique.  Of the 18.8K entities that exist in KVQA many are rare; only 65\% of the entities exist in the top 1 million most occurring entities in Wikipedia and only 91\% of them were found in WikipediaVec's entity matrix.

The baseline model proposed in the KVQA dataset paper is composed of both visual and text based entity linking methods to Wikidata from which entity facts over 18 pre-determined relationship types are extracted. These facts are then encoded along with spatial coordinates and text questions via memNet or Bi-LSTM and fed into a multi-layer perceptron followed by a softmax classifier to predict the final answer from among all 20k possible answers. 

We note that around 25\% of the images in the main dataset, 6K out of 24K, are found directly in the 69K image reference dataset for face identification making visual entity linking relatively simple and unrealistic.  Additionally the model only provides results for a closed world experiment where only 18 types of relations are considered from 18.8k Wiki entity pages.  In contrast our method does not rely on a reference set of images for entity identification nor a subset of relations for use as answer possibilities.

\paragraph{KVQA entity span construction}
KVQA does not provide gold entity spans and we examine three methods for generating them, the first two which involve using the SpaCY named entity recognizer \footnote{We use ``web-sm" from http://spacy.io}. In total, we have 5 different variants of the data we try in our experiments as input where the first two ``Question" and ``+ Caption" are baselines which do not utilize knowledge injection.
\begin{enumerate}
    \item ``Question" - only the question is used
    \item ``+ Caption", the question with the image caption is used
    \item ``NERper" uses SpaCY to first identify named entity spans and then filters to only include those whose labels are associated with people.  
    \item ``NERagro" uses SpaCY with an additional noun phrase chunking mechanism and no filtering to aggressively catch more spans.  
    \item ``KVQAmeta" uses the wiki entity names provided in the metadata of each question and uses string matching to search for them over the question.  In cases where entities are not present in the question, we prepend them to the caption.  
\end{enumerate} 
The KVQAmeta procedure allows for the possibility of knowledge injection of all entities present in a question, while NERper and NERagro allow for additional entities that might be present in the question and image caption to be introduced.  In all three cases the image caption is concatenated to the end of the question prior to running the entity span detection methods.  For these three sets, we also try three methods to improve entity linking.
\begin{enumerate}
    \item ``as is" which utilize the entity sets as is, 
    \item ``links" which filters the entity sets to only include those with verified links to wikipedia and 
    \item ``noisy" which leverages a Wikipedia API\footnote{https://github.com/goldsmith/Wikipedia} to search for the most likely wikipedia link for spans missing them. 
\end{enumerate}
Table \ref{table:kvqa_overall_table} shows the entity spans per question, E-BERT injected entities per question (since only those with values in WikipediaVec can be mapped) and the percent of questions with E-BERT injections for split 1.  We note that KVQAmeta has the least E-BERT injected entities per question and the highest percent of questions with E-BERT injected entities ( 99\% for KVQAmeta "noisy" for instance), while NERagro has the most E-BERT injected entities per question and NERper has the lowest percent per question.
\begin{table}
  \centering
\begin{tabular}{l c c c c c}
\toprule
   &  &   & ents  & eberts & Qs w/ \\
Model & Type & Acc & per Q &  per Q & eberts\\
\midrule
  Shah 2019 & - & 49.50 & - & - & -\\
  + Caption & - & 50.20 & - & - & - \\
\midrule
  Question &   -   &    47.54 & - & - & - \\
  + Caption &    -  &    50.25 & - & - & - \\
\midrule
 NERper & as is &         50.37 &       2.5 &   1.5  &     .78 \\          
 NERper & links &         50.42 &       1.8 &   1.5  &     .79 \\         
 NERper & noisy &         50.69 &       2.5 &   2.3  &     .94 \\         
\midrule                               
 NERagro & as is &        50.26 &  \textbf{4.0} &   2.6  & .91 \\
 NERagro & links &        50.33 &       2.2 &   2.2  &     .97 \\      
 NERagro & noisy &        50.77 &       3.3 &   \textbf{3.2}  & .97 \\  
\midrule                                    
 KVQAmeta & as is &          52.65 &       1.4 &   1.2  &     .87 \\    
 KVQAmeta & links &          52.68 &       1.4 &   1.3  &     .95 \\    
 KVQAmeta & noisy & \textbf{52.83} &       1.4 &   1.4  & \textbf{.99} \\
\bottomrule
\end{tabular}
\caption{KVQA overall accuracy results over 5 splits and entity spans per question (ents per Q), E-BERT representations injected per question (eberts per Q) and the percent of questions with E-BERT injections (Qs w/ eberts) for split 1}
\label{table:kvqa_overall_table}
\end{table}

\paragraph{OKVQA dataset} The OKVQA dataset \cite{okvqa} from AllenNLP contains 14k image/question pairs which is less than KVQA and is less entity based since its objective is to test commonsense reasoning. It does however provided around 10 answers per question which is more robust from an answer set perspective than KVQA which only includes one labeler's answer per question since slight variations of a labeler's answer would cause incorrect answers. 

\paragraph{OKVQA entity span construction}  As OKVQA does not provide entity spans, we again leverage SpaCY to get 3 different entity span set variants. These three sets represent progressively less noisy versions of entity sets obtained which we hypothesized would be beneficial given OKVQA is less entity-centric overall compared with KVQA.
\begin{enumerate}
    \item ``13K" we use SpaCY with no filtering to obtain entity spans for 13K QA pairs (92.8\% of questions).
    \item ``4K" we use a semi-automated rules based technique to identify poor candidate spans (ie too general, etc) which filters the set to ``4K" (28.6\% of questions). 
    \item ``2.5K" we did manual filtering over unique entity spans to filter it down to ``2.5K" (17.8\% of questions).  
\end{enumerate}

\begin{table}
  \centering
\begin{tabular}{l c c c c}
\toprule
  Model &  Mean &  Std  &  Max  &  Median \\
\midrule
OKVQA best & 27.84 & - & - & -\\
Shevchenko 21 & 39.04 & - & - & -\\
Wu et al 21 & 40.50 & - & - & -\\
\midrule
  LXMERT Plain & 43.51 & 0.23  & 43.87 & 43.34 \\
  + EBERT 13K & 40.59 & 0.09  & 40.69 & 40.59\\
  + EBERT 4K & \textbf{43.67} & 0.13  & 43.88 & \textbf{43.66}\\
  + EBERT 2.5K & 43.61 & 0.36  & \textbf{44.10} & 43.34\\
\bottomrule
\end{tabular}
  \caption{OKVQA model results over 5 runs}
  \label{table:okvqa_table}
\end{table}

\begin{table*}
  \centering
  
\begin{tabular}{ll rrrrrrrrrr c}
\toprule
& & & multi & multi & & multi & & & & & & Acc /\\
Model & Type &  1-hop &  hop &  rel &  bool &  entity &  cmp &  spatial &  subtr &  count &  inter &   Conf \\
\midrule
Percent &     with &  81.80 &      18.20 &           53.58 &    24.63 &         24.96 &       16.81 &    15.22 &        12.07 &      7.89 &          1.82 &  -\\
\midrule  
Question &        - &  44.89 &      57.98 &           47.40 &    86.37 &         72.14 &       81.67 &    28.12 &        19.68 &     84.62 &         65.00 & 47.27 \\
+ Caption &        - &  46.36 &      65.47 &           51.57 & \textbf{87.21} &         72.46 &       80.91 &    29.17 &        19.33 &     85.03 &         70.29 & 49.84 \\
KVQAmeta &    links &  48.87 &      70.61 &           55.43 &    86.69 & \textbf{73.68} & \textbf{82.50} &    31.14 & \textbf{22.21} &     84.82 & \textbf{71.47} & 52.83 \\
KVQAmeta &    noisy &  \textbf{48.88} & \textbf{71.55} & \textbf{56.14} &    86.63 &         73.57 &       82.15 &    31.14 &        21.23 &     \textbf{85.70} &         70.00 & \textbf{53.01} \\
\midrule
Average & E-BERT & 47.38	& 67.48	& 53.04	& 86.24	& 72.98	& 81.85	& 30.48	& 20.58	& 85.15 &	68.46 & 51.04 \\
Best E-BERT& - Caption & 2.52 & 6.08 & 4.57 & -0.13 & 1.22 & 1.59 & 2.25 & 2.88 & 0.67 & 1.18 & 3.17 \\
\midrule
 Question &     - &  -0.01 &       1.32 &            0.05 &     3.20 &          2.21 &        2.89 &    -1.69 &        -1.79 &      5.57 &          1.76 &  0.23  \\
  + Caption &     - &   0.50 &       2.70 &            1.00 &     4.26 &          3.15 &        3.85 &    -1.18 &        -1.83 &      5.97 &          3.52 &  0.90  \\
  KVQAmeta & links &   1.08 &       4.26 &            1.99 &     4.65 &          3.54 &        4.16 &    -0.71 &        -1.52 &      6.86 &          3.54 &  1.66  \\
  KVQAmeta & noisy &   \textbf{1.52} & \textbf{4.84} & \textbf{2.48} & \textbf{5.87} & \textbf{4.34} & \textbf{5.02} & \textbf{-0.44} & \textbf{-1.51} & \textbf{7.31} & \textbf{5.24} & \textbf{2.12}  \\
\bottomrule
\end{tabular}
  \caption{KVQA results by question type accuracy (top half) and confidence ( bottom 4 rows of unconstrained logits).  Not shown NERper has highest accuracy for spatial question types (31.42).  Average E-BERT refers to averages over NERper, NERagro and KVQAmeta for each link type (as is, links, noisy) }
  \label{table:kvqa_split1_qt_conf_table}
\end{table*}

\section{Results}

Table \ref{table:kvqa_overall_table} shows the average results over the 5 KVQA splits\footnote{See Appendix Table \ref{table:kvqa_res_table} for per split results} and Table \ref{table:okvqa_table} shows the results over 5 runs with random seeds for OKVQA.  In following we highlight observations of interest as they pertain to knowledge injection for VQA.

\paragraph{How E-BERT effects task accuracy}  For KVQA, we see the KVQAmeta noisy entity set based model provides the best results ( 52.83 ) compared with the results of feeding the same question + caption text into LXMERT without knowledge injection ( 50.25 ).  Additionally using E-BERT on the NERper and NERagro noisy search entity spans gives 0.5 accuracy improvement confirming the utility and efficiency of our method for KBVQA in those cases as well. 

The importance of entity span quality is evidenced by the variation in results between NERper, NERagro and KVQAmeta and in all 3 cases using the "noisy" search mechanism to find entity links to Wikipedia provides the best results.  We finally note that using LXMERT with questions + captions outperforms the KVQA paper's baseline model \cite{shahMYP19} which relies on a closed world subset of Wikipedia setting that only considers facts from 18 relations as candidate answers and a simplified face identification entity linking step,  none of which are used in our setup.

For OKVQA, we see that adding E-BERT to LXMERT only slightly improves results compared with using LXMERT without knowledge injection and only when the entity sets provided ( 4K and 2.5K ) are less noisy than those provided by SpACY outright (13K). The OKVQA data is less entity centric and does not contain image captions so retrieval or generation methods for captions could be useful.

We note that LXMERT Plain, which does not use knowledge injection, already does better than the results from the OKVQA paper baseline model \cite{okvqa} (43.51 vs 27.84) and the two highest performing models  \cite{shevchenko2021reasoning,jialin} when our experiments were run\footnote{https://okvqa.allenai.org/leaderboard.html}.  The authors in \cite{shevchenko2021reasoning} redo the costly step of pre-training of LXMERT with their form of knowledge injection and get 39.04 accuracy over 3 runs whereas our technique does not require rerunning pre-training. As noted in \cite{jialin}, the OKVQA test images are a subset of COCO validation images which are used to pre-train most of transformer-based vision and language models including LXMERT and VilBERT \cite{lu2019vilbert}. Although the test questions never appear in the pre-training process, other questions on the test images may help the system understand the image better, leading to a higher performance.  In \cite{jialin} that amounts to a 1.1 accuracy difference ( from 40.5 to 39.4) when redoing pretraining ViLBERT with the OKVQA test images that appear in COCO removed.   At the time when our experiments were run we obtained the state of the art for OKVQA though with the aforementioned data leakage pre-training issue. 

\paragraph{When E-BERT effects task accuracy}  To better understand when E-BERT knowledge effects task accuracy, we show accuracy and confidence results by question type for split 1 of KVQA in Tables \ref{table:kvqa_split1_qt_conf_table}. We see all models perform poorly at questions of type ``subtraction" and ``spatial"\footnote{See Appendix Figures \ref{fig:kvqa_qs_ss_1} and \ref{fig:kvqa_qs_ss_2} for examples of question types.} which represent 12\% and 15\%, of questions with combined  E-BERT average accuracies of 20.6 and 30.5.  We see that both types of questions are quite challenging and almost entirely based on visual entity identification. Adding image captions to the question gives a slight improvement in ``spatial" questions and worse performance in ``subtraction" ones. In both cases E-BERT results using ``NERper noisy" and ``KVQAmeta links" give 2.2 and 2.8 point improvements. 

For question types where LXMERT already performs strongly, ``boolean", ``counting" and ``comparison" with average accuracies of 86.3, 85.1 and 81.8, we see the improvements provided via the best E-BERT results ( -.01, .7 and 1.6 ) are generally smaller than those for other question types such as ``multi-hop", ``multi-relation", and ``subtraction" which get improvements of 6.1, 4.6 and 2.9 points. We see that across question types the models that utilize E-BERT are more confident than the models which do not use knowledge injection though the level of over-confidence is inline with widely known calibration issues affecting neural nets. \citep{desai-durrett-2020-calibration, Jiang2021HowCW}

\paragraph{E-BERTs effect on VQA explainability} We extract visual and text explanations using BM-GAE and TRF on our KVQA models and Table \ref{table:kvqa_expl_bgm_trf_table} shows accuracy results for when these explanation methods find E-BERT enhanced entities to be in the top 5 most important tokens leading to a given answer prediction\footnote{Appendix Table \ref{table_kvqa_expl_table} shows the percent breakdown of how many entities appear in the top 1, top 5 and top 10 important tokens for each explanation type over our entity sets along with the percent of questions with E-BERT entity injection for each.}.  We see that for 7 of the 9 models, questions which include E-BERT entities amongst their top 5 using BM-GAE provide better accuracy than those using the TRF method.  Averaging over the models, we'd achieve 59.74\% accuracy with the BM-GAE model and 58.33\% using the TRF method compared with 51.04\% average accuracy amongst all E-BERT injection models as seen in Table \ref{table:kvqa_split1_qt_conf_table}.  This finding suggests that when using either method, an entity appearing in the top 5 most important tokens allows for improved accuracy which is in agreement with the perturbation testing results in \citep{BMGAE}. 

Over all models E-BERT entities appear in the top 5 most important tokens using TRF more than BM-GAE (10.35 vs 8.59 \%) though for 3 of the models, the BM-GAE method finds more. 
Interestingly using the explaination methods on the KVQAmeta noisy entity sets model, which obtains the best task accuracy results, leads to worse results compared with using the ``as is" and links version of the KVQAmeta entity set which suggest that for these questions the ``noisy" use of the wikipedia search API to link possible entity spans to entity pages had an adverse effect which is not present when considering all questions, as opposed to just those where E-BERT entities are in the top 5 important tokens.  

We explore qualitative trends in these explanations to see where knowledge injection is helpful, specifically where the KVQAmeta model predicts correctly while the ``+ Caption" does not. \footnote{Two such examples are found in Appendix Figures \ref{fig:kvqa_bmgae_trf_q1} and \ref{fig:kvqa_bmgae_trf_q2}}. In these cases injecting entity knowledge focuses the model on these entities which is evidenced by the top tokens shown for the KVQAmeta model.  In the first case, only KVQAmeta injects ``Knute Nelson" which appears as the 2nd most important token and leads to the only correct prediction (Europe) amongst the models.  The token explanations show knowledge injection lessens the importance of question specific words like ``which" and``how", which are in top 5 tokens for the Question only model. This behavior makes sense since LXMERT was not trained with captions ( knowledge injected or not ) and fine tuning with this extra context shifts the domain a bit.  

We also explore examples where E-BERT hurts model performance\footnote{See Appendix Figures \ref{fig:kvqa_bmgae_trf_q1b} and \ref{fig:kvqa_bmgae_trf_q2b} for negative examples}.  In the first example, while the Question only and ``+ Caption" models answer correctly, the entity``Top Gun" is injected for the NERper and NERagro models leading the model to incorrectly predict the entity ``Duke Cunningham" was born after WW2, possibly due to Top Gun the 1986 film, and not the US Navy school, being erroneously injected. In the KVQAmeta model, injecting the true entity ``Duke Cunningham" causes the model to change its prediction incorrectly which reflects a true error use case.  The second example on the other hand shows a label error as the entity Fisher Morgan\footnote{https://en.wikipedia.org/wiki/Fisher\_Morgan} is in fact both a singer and actor, the later which KVQAmeta is actually the only model to predict correctly.  A way to identify similar dataset label errors would be to inspecting test cases where KVQAmeta injects entities not found by the other E-BERT models and gives a unique prediction marked as incorrect.

\begin{table}
  \centering
\begin{tabular}{ll rr rr}
\toprule
          & & BM & BM & TRF & TRF \\
          Model & Type &  ACC & Qs &  Acc & Qs\\
\midrule
  NERper & as is &    \textbf{58.25} &   11.48 &    56.11 &    6.13 \\
  NERper & links &    \textbf{62.18} &    8.67 &    56.28 &    6.90 \\
  NERper & noisy &    \textbf{69.85} &    4.75 &    68.17 &    7.11 \\
\midrule
 NERagro & as is &    \textbf{65.91} &    4.93 &    62.41 &    7.41 \\
 NERagro & links &    \textbf{52.74} &   14.75 &    49.31 &   18.52 \\
 NERagro & noisy &    \textbf{56.07} &   20.53 &    43.31 &   18.23 \\
\midrule
KVQAmeta & as is &    61.00 &    2.77 &    \textbf{70.03} &    6.30 \\
KVQAmeta & links &    68.97 &    4.26 &    \textbf{79.67} &   12.57 \\
KVQAmeta & noisy &    \textbf{42.72} &    5.15 &    39.65 &   10.02 \\
\midrule
 Average &       &    \textbf{59.74} &    8.59 &    58.33 &   10.35 \\
\bottomrule
\end{tabular}

  \caption{KVQA Bi-modal (BM) and Transformer attention (TRF) explaination results for Questions where an E-BERT injected entity is in top 5 most important tokens.}
  \label{table:kvqa_expl_bgm_trf_table}
\end{table}

\paragraph{Effect of additional types of knowledge injection} In an effort to see whether other retrieval augmented, nearest neighbor or confidence based methods, all applied without redoing pre-training,  would improve upon the results of our knowledge enhanced model using the ``KVQAmeta" derived entity sets for the KVQA data, we conducted initial experiments using Dense Passage Retrieval (DPR) for Open Domain QA \cite{karpukhin-etal-2020-dense}, kNN language models (knn LMs) \cite{khandelwal2020generalization, naz} and simple confidence thresholding where E-BERT was only utilized when the model's confidence was above a given threshold determined on a hold out validation set.  Although we do see slight improvements with the final confidence method, all within a point accuracy, we note that retrieving additional text ( 3 lines ) per question with DPR or using nearest neighbor semantic similarity lookup (knn LMs) over the training set of the downstream task did not lead to any sizeable improvement ( usually within 0.1 accuracy ).  It seems likely that in both the later cases, these methods would need to redo pre-training of LXMERT in order to see gains and as our focus is on studying the effects of simple, efficient knowledge injection during fine-tuning we leave that as future work.

\section{Conclusion}
\indent In this work we analyze how efficient knowledge injection via E-BERT applied during fine tuning affects the performance of an existing visual-linguistic model LXMERT on the relatively unexplored task of knowledge-based VQA (KBVQA) both in terms of accuracy and explainability via BM-GAE.  We experiment using two large publicly available VQA datasets: (i)KVQA \citep{shahMYP19} that is explicitly tied to Wikipedia and rich in rare entities and (ii) OKVQA \citep{okvqa} that is less entity-centric and more aligned with common sense reasoning. Both datasets lack explicit entity spans and we show how using different entity sets resulting from either weakly supervised methods or manual human annotation affects knowledge injection on task performance.  Our analysis shows improved performance on the entity rich KVQA data, 2.5\% top 1 accuracy, and a smaller improvement on OKVQA, both without the need to redo any costly pre-training.

In the future we can study how and if such knowledge injection techniques benefit other VQA models including recent work on prompt tuning GPT3\citep{yang2021empirical} which shows improved results on OKVQA, though there the model size, in terms of model parameters, increases from 228 million for LXMERT to 175 billion for GPT3.  The authors use of VinVl \citep{zhang2021vinvl} and COCO for generated and gold image captions could improve our results on OKVQA.

\bibliography{aaai22.bib}  

\appendix
\section{Appendix}

\begin{figure*}[t]
    \centering
    \includegraphics[width=1\linewidth]{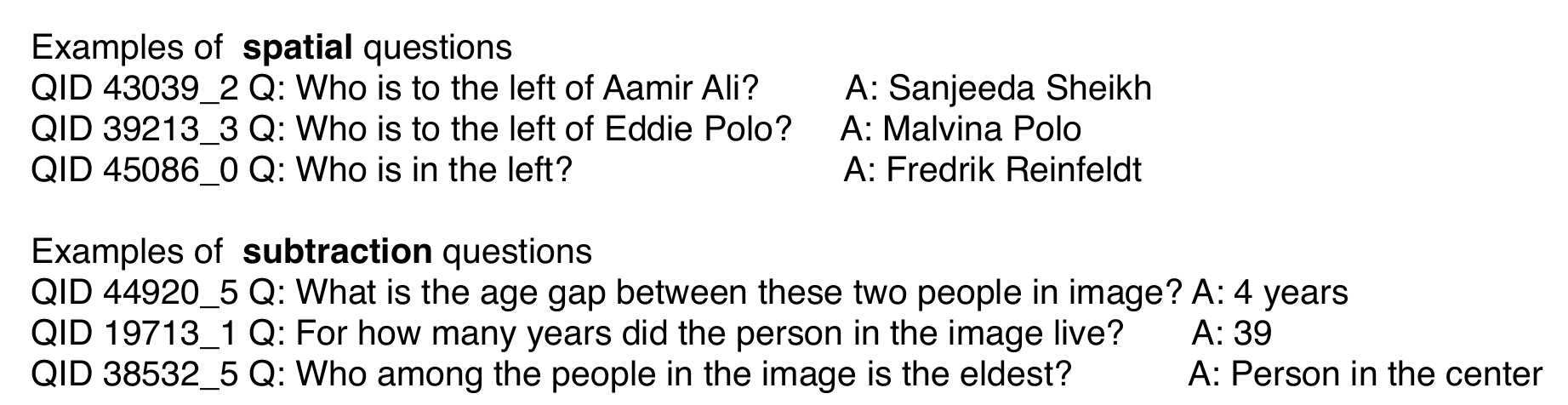}
    \caption{Examples of KVQA questions by question type where all models perform poorly}
    \label{fig:kvqa_qs_ss_1}
\end{figure*}

\begin{table*}
  \centering

\begin{tabular}{ll|rrrrr|r|rrrr}
\hline
  Model &  Type &  S 1 &  S 2 &  S 3 &  S 4 &  S 5 &   Avg & Std & Max & Median & \% $>$ Capt\\
\hline
 \# questions & test & 18697 & 18505 & 18696 & 19248 & 18120 & & & & & \\
\hline
  Plain &     - &    47.27 &    47.54 &    47.34 &    48.35 &    47.18 & 47.54                                                     &  0.47 &  48.35 &   47.34 & 0\\
   Capt &     - &    49.84 &    50.43 &    50.66 &    51.05 &    49.25 & 50.25                                                     &  0.71 &  51.05 &   50.43 & 0\\
 \hline  
 NERper & as is &    50.05 &    50.67 &    50.25 &    51.58 &    49.28 & 50.37                                                     &  0.85 &  51.58 &   50.25 &  0.8\\
 NERper & links &    50.11 &    50.28 &    50.73 &    50.27 &    50.73 & 50.42                                                     &  0.29 &  50.73 &   50.28 &  0.6\\
 NERper & noisy &    51.10 &    50.24 &    50.25 &    51.34 &    50.53 & 50.69                                                     &  0.50 &  51.34 &   50.53 &  0.6\\
 \hline
 NERagro & as is &    49.72 &    50.05 &    50.33 &    51.09 &    50.11 & 50.26                                                    &  0.51 &  51.09 &   50.11 &  0.4\\
 NERagro & links &    50.29 &    50.24 &    49.64 &    50.86 &    50.63 & 50.33                                                    &  0.46 &  50.86 &   50.29 &  0.4\\
 NERagro & noisy &    50.08 &    50.78 &    50.72 &    51.38 &    50.91 & 50.77                                                    &  0.47 &  51.38 &   50.78 &  1.0\\
 \hline
 KVQAmeta & as is &    52.20 &    52.64 &    \textbf{52.70} &    53.22 &    52.50 & 52.65                                          &  0.37 &  53.22 &   52.64 &  1.0\\
 KVQAmeta & links &    52.83 &    52.60 &    51.92 &    53.44 &    52.60 & 52.68                                                   &  0.55 &  53.44 &   52.60 &  1.0\\
 KVQAmeta & noisy &    \textbf{53.01} &    \textbf{52.86} &    52.37 &    \textbf{53.58} &    \textbf{52.34} & \textbf{52.83}      &  0.51 &  53.58 &   52.86 &  1.0\\
\hline
\end{tabular}

\caption{KVQA results over all splits, models and link types. Last column is Percent of Times model outperforms CAPT results}
\label{table:kvqa_res_table}
\end{table*}

\begin{table*}

  \centering
\begin{tabular}{ll|lll|lll|c}
\hline
& & \multicolumn{3}{c}{bimodal generic} & \multicolumn{3}{c}{transformer attention}  & Qs w/\\
 Model & Type &  top1 &  top5 &  top10 &  top1 &  top5 &  top10 & EBERT\\
\hline
NERper & as is & 0.66 & 11.48 & 31.23 & 0.29 & 6.13 & 22.64 & .78 \\
NERper & links & 0.32 & 8.67 & 33.32 & 0.39 & 6.90 & 25.24 & .79\\
NERper & noisy & 0.13 & 4.75 & 21.62 & 0.73 & 7.11 & 23.38  & .94\\
\hline
NERagro & as is & 0.31 & 4.93 & 19.60 & 0.38 & 7.41 & 28.32 & .91\\
NERagro & links & 0.56 & 14.75 & 44.46 & 1.10 & 18.52 & 50.02 & .97\\
NERagro & noisy & 1.30 & 20.53 & 44.94 & 1.43 & 18.23 & 40.95 & .97 \\
\hline
KVQAmeta & as is & 0.12 & 2.77 & 8.52 & 0.18 & 6.30 & 15.56 & .87 \\
KVQAmeta & links & 0.39 & 4.26 & 12.96 & 4.06 & 12.57 & 23.80 & .95 \\
KVQAmeta & noisy & 0.15 & 5.15 & 23.75 & 0.42 & 10.02 & 36.19 & .99 \\
\hline
\end{tabular}
  \caption{KVQA entity knowledge injection explainability on split 1 for various entity span sets.  For instance, 11.48 \% of inference questions have E-BERT entities in their top 5 tokens for the NERper plain entity set model and overall 78\% of questions in that entity set have E-BERT injected entities.}
  \label{table_kvqa_expl_table}
\end{table*}

\begin{figure*}[t]
    \centering
    \includegraphics[width=1\linewidth]{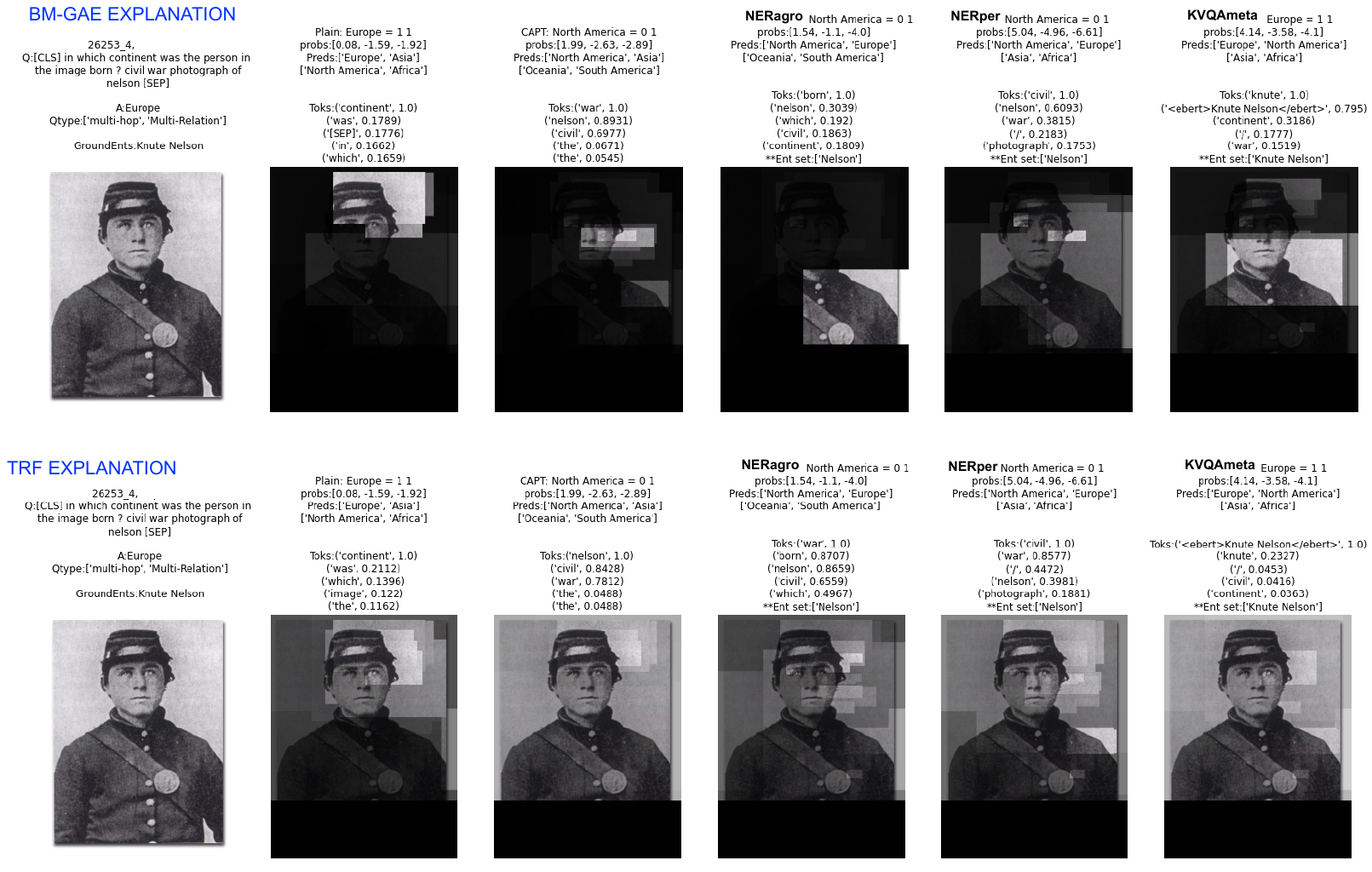}
    \caption{Example 1 of a KVQA question where E-BERT is beneficial for KVQAmeta noisy entity set model. The rows show visual and token explanations for BM-GAE and TRF over the question/text (left column) and the 5 variants ``Question", ``+Caption", NERagro, NERper and KVQAmeta we explore .  Next to each models name is their prediction and whether this top1 prediction is correct (1) or not, and then whether the correct answer exists in the top 5 predictions of the model which are additionally shown along with their logit values. Below that we see the top 5 most important tokens found by the explanation method followed by the set of Entities used for possible knowledge injection.}
    \label{fig:kvqa_bmgae_trf_q1}
\end{figure*}

\begin{figure*}[t]
    \centering
    \includegraphics[width=1\linewidth]{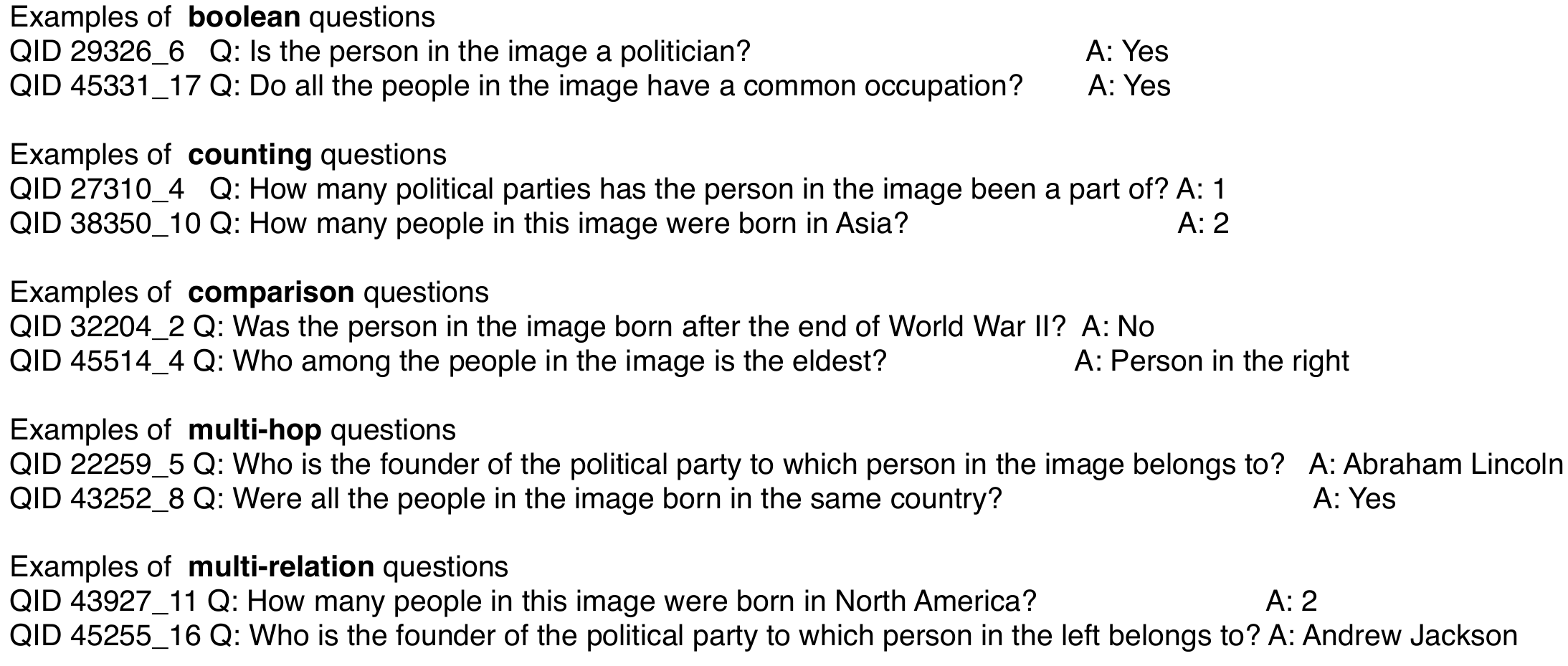}
    \caption{More examples of KVQA questions by question type}
    \label{fig:kvqa_qs_ss_2}
\end{figure*}

\begin{figure*}[t]
    \centering
    \includegraphics[width=.95\linewidth]{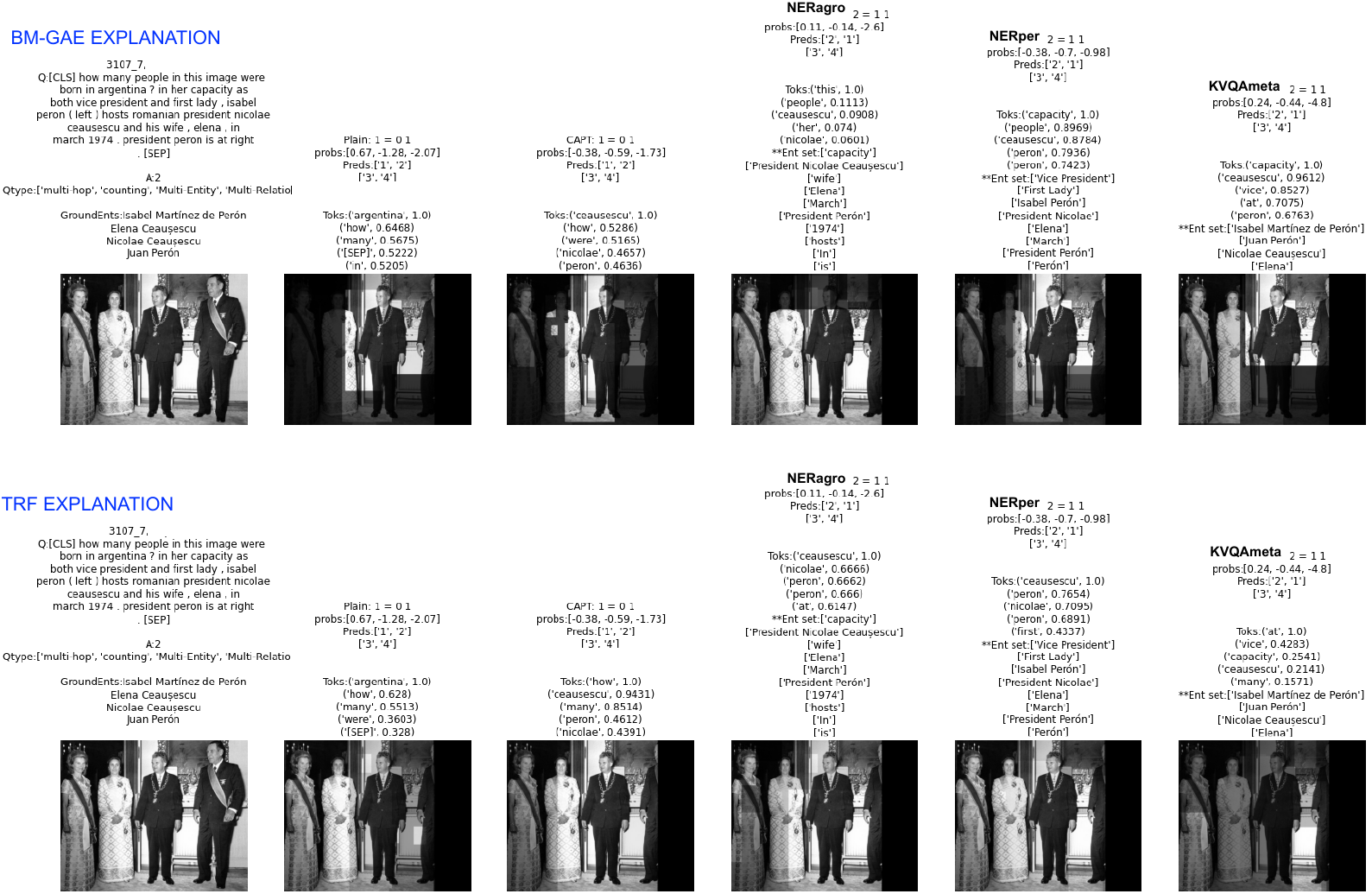}
    \caption{Example 2 of BM-GAE and TRF explanations for a KVQA question where E-BERT is beneficial for KVQAmeta}
    \label{fig:kvqa_bmgae_trf_q2}
\end{figure*}

\begin{figure*}[t]
    \centering
    \includegraphics[width=.95\linewidth]{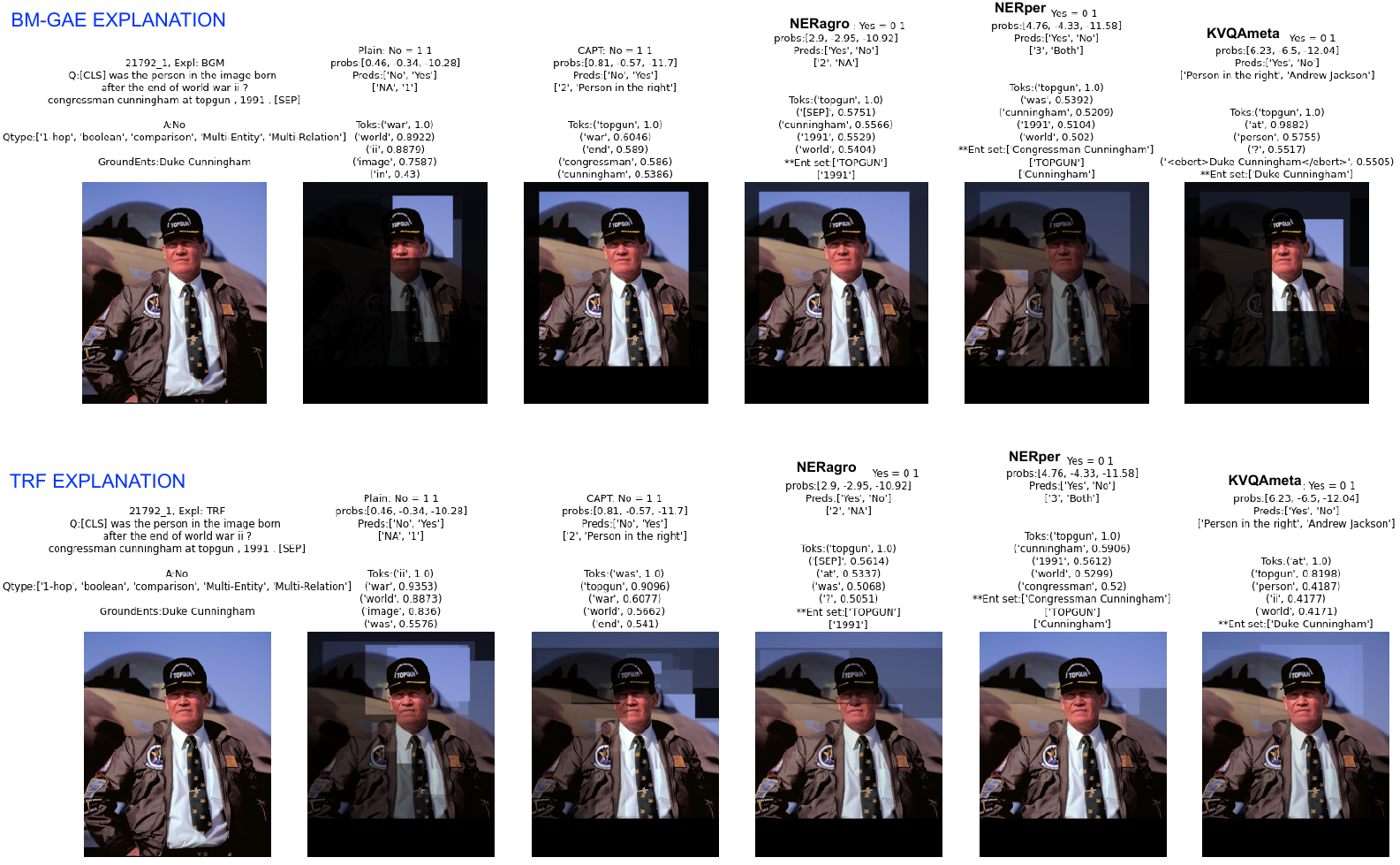}
    \caption{Example 1 of BM-GAE and TRF explanations for a KVQA question where E-BERT is harmful for KVQAmeta}
    \label{fig:kvqa_bmgae_trf_q1b}
\end{figure*}

\begin{figure*}[t]
    \centering
    \includegraphics[width=.95\linewidth]{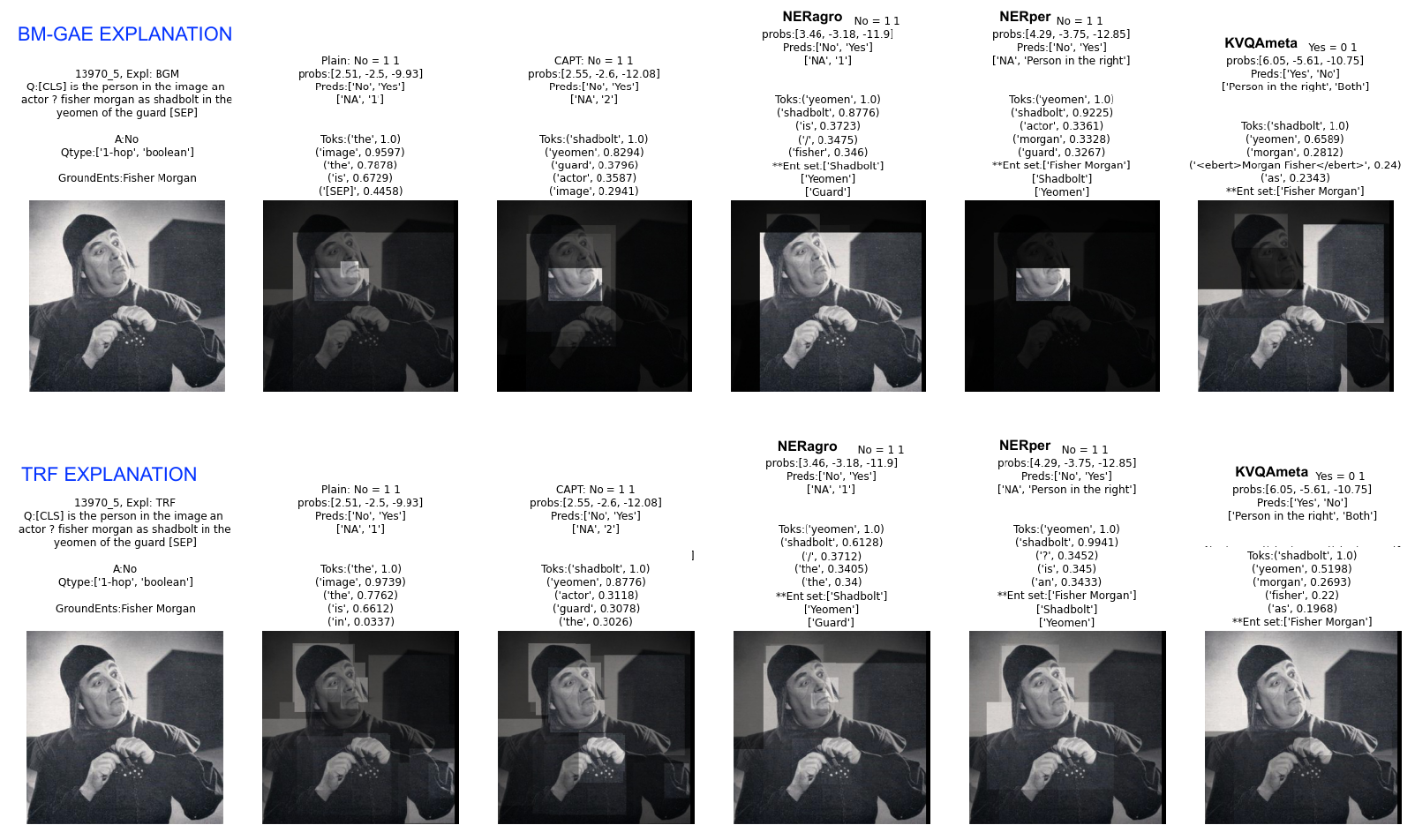}
    \caption{Example 2 of BM-GAE and TRF explanations for a KVQA question where E-BERT is harmful for KVQAmeta}
    \label{fig:kvqa_bmgae_trf_q2b}
\end{figure*}

\end{document}